\documentclass[10pt,twocolumn,letterpaper]{article}
\usepackage{caption, subcaption, multirow, overpic, textpos}
\usepackage{iccv}
\usepackage{times}
\usepackage{epsfig}
\usepackage{graphicx}
\usepackage{amsmath}
\usepackage{amssymb}
\usepackage{booktabs}
\usepackage{algorithm, algorithmic}
\usepackage{multirow}

\usepackage[pagebackref=true,breaklinks=true,letterpaper=true,colorlinks,bookmarks=false]{hyperref}

\usepackage[capitalize]{cleveref}
\crefname{section}{Sec.}{Secs.}
\Crefname{section}{Section}{Sections}
\Crefname{table}{Table}{Tables}
\crefname{table}{Tab.}{Tabs.}

\iccvfinalcopy % *** Uncomment this line for the final submission

\begin{document}

%%%%%%%%% TITLE
\title{Latent Feature Relation Consistency for Adversarial Robustness}

\author{
Xingbin Liu$^1$ \quad Huafeng Kuang$^1$ \quad Hong Liu$^1$ \quad Xianming Lin$^1$ \quad Yongjian Wu$^2$ \quad Rongrong Ji$^{1}$
\\
$^{1}$ Media Analytics and Computing Lab, Department of Artificial Intelligence,\\
School of Informatics, Xiamen University, 361005, China.\\
$^2$ Tencent Youtu Lab, Shanghai, China\\
}

\maketitle  
% Remove page # from the first page of camera-ready.
% \ificcvfinal\thispagestyle{empty}\fi

%%%%%%%%% ABSTRACT
\begin{abstract}
Deep neural networks have been applied in many computer vision tasks and achieved state-of-the-art performance.
However, misclassification will occur when DNN predicts adversarial examples which add human-imperceptible adversarial noise to natural examples.
This limits the application of DNN in security-critical fields.
To alleviate this problem, we first conducted an empirical analysis of the latent features of both adversarial and natural examples and found the similarity matrix of natural examples is more compact than those of adversarial examples.
Motivated by this observation, we propose \textbf{L}atent \textbf{F}eature \textbf{R}elation \textbf{C}onsistency (\textbf{LFRC}), which constrains the relation of adversarial examples in latent space to be consistent with the natural examples.
Importantly, our LFRC is orthogonal to the previous method and can be easily combined with them to achieve further improvement. 
To demonstrate the effectiveness of LFRC, we conduct extensive experiments using different neural networks on benchmark datasets. 
For instance, LFRC can bring 0.78\% further improvement compared to AT, and 1.09\% improvement compared to TRADES, against AutoAttack on CIFAR10.
Code is available at \url{https://github.com/liuxingbin/LFRC}.
\end{abstract}

%%%%%%%%% BODY TEXT
\section{Introduction}
Deep Neural Networks (DNNs) have achieved state-of-the-art performance in various vision tasks, such as segmentation~\cite{pal1993review}, detection~\cite{redmon2016you}, and super-resolution~\cite{glasner2009super}, which are based on the development of fancy algorithms and powerful computational resources like modern GPUs.
However, DNNs are extremely vulnerable to adversarial perturbations which are humanly imperceptible.
Adversarial attacks covered many deep learning tasks including classification~\cite{goodfellow2014explaining}, person re-identification~\cite{wang2020transferable}, and natural language processing~\cite{li2020bert}. 
Considering the wide range of applications of neural networks in our world, especially in safety-critical areas such as autonomous driving~\cite{chen2019model} and medical diagnosis~\cite{kong2017cancer}, it is urgent to improve the robustness of neural networks.

%--------framework--------------------
\begin{figure}
    \begin{center}
    \includegraphics[width=\columnwidth]{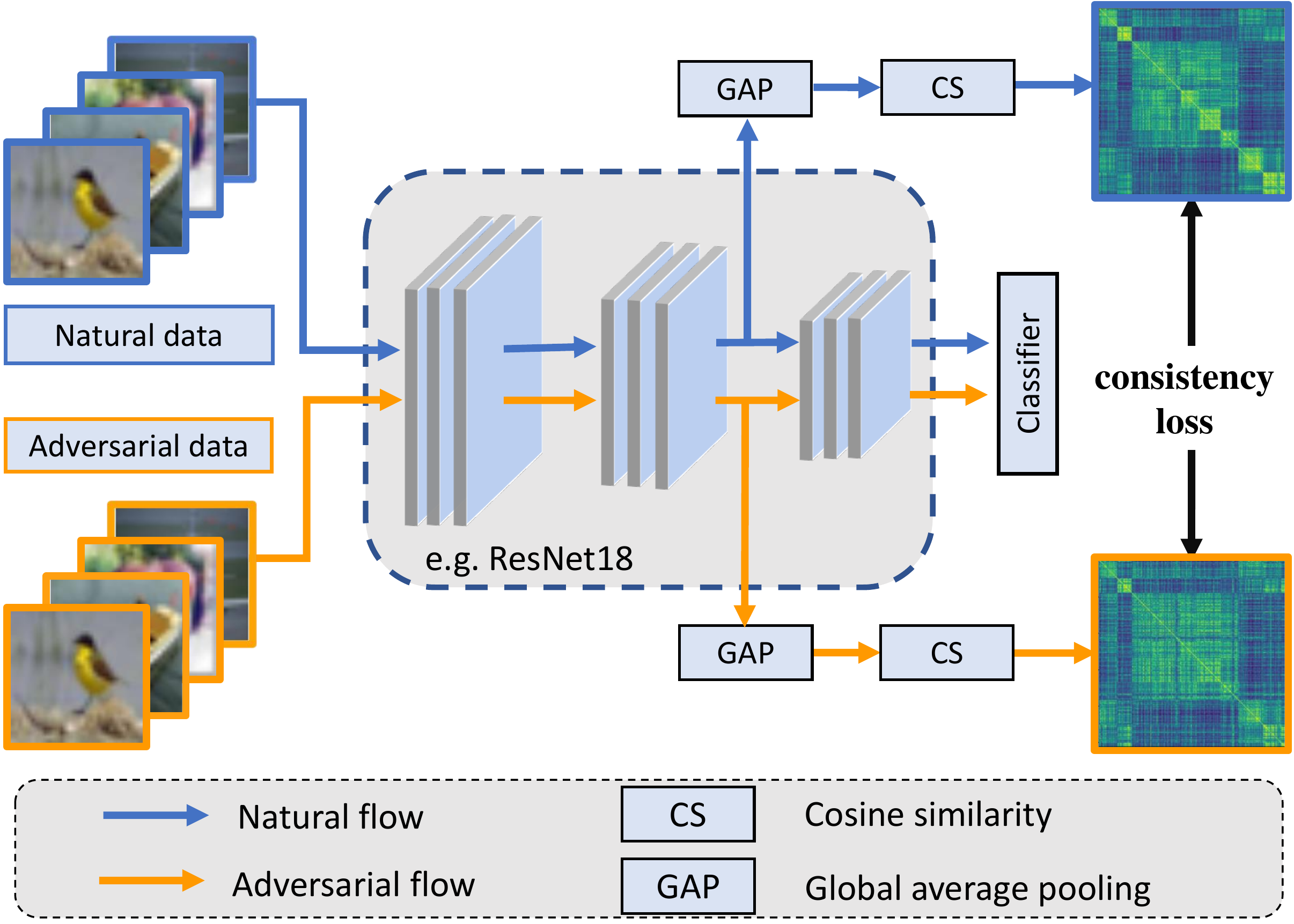}
    \end{center}
    \caption{The framework of the proposed method (LFRC). During training, the natural data and adversarial data are drawn from the natural data space and the adversarial data space respectively. We feed-forward the natural and adversarial examples to the neural network to obtain their latent features. Subsequently, the latent features are passed through a global average pooling (GAP) layer, and then the cosine similarity is calculated to obtain the similarity matrix. Finally, we constrain the similarity matrices of the adversarial and natural examples.}
    \label{framework}
\end{figure}

\begin{figure*}
  \begin{center}
  \includegraphics[width=1.0\textwidth]{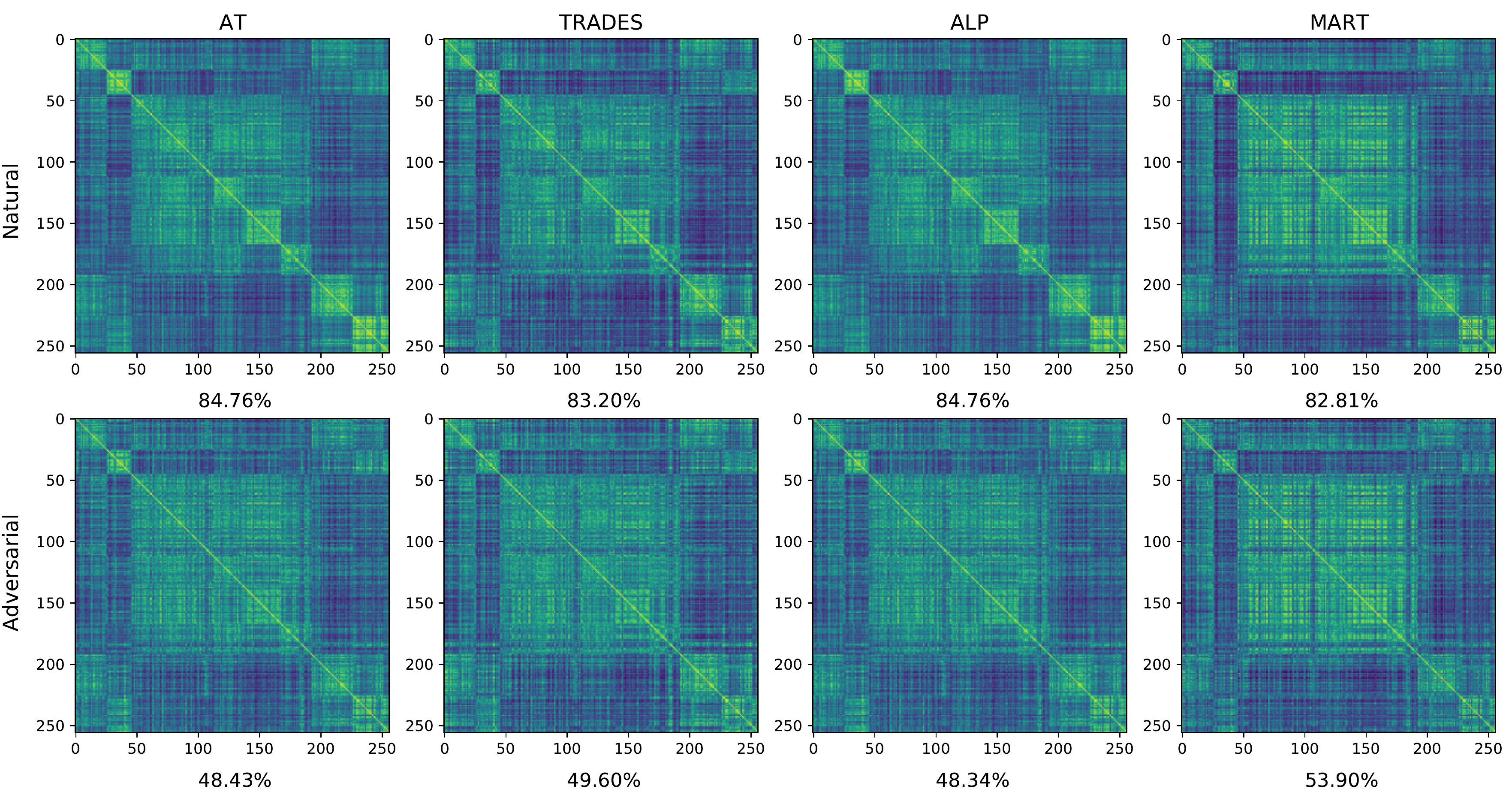}
  \end{center}
  \caption{Similarity matrices of the adversarial and natural examples on the first batch of CIFAR10 test set for the ResNet18 network trained by different methods. The examples in each batch have been grouped by their true label for better visualization(grouping doesn't been operated during the train and test procedure). The first row is the similarity matrix of natural examples and the second row is the similarity matrix of adversarial examples. From the left column to the right column are the results of AT~\cite{madry2017towards}, TRADES~\cite{zhang2019theoretically}, ALP~\cite{kannan2018adversarial}, and MART~\cite{wang2019improving}, respectively. The accuracy(\%) of this batch is noted at the bottom of each figure. Compared with the feature similarity matrix of adversarial examples, distinct bright blocks can be found on the diagonal of the similarity matrix of natural examples, corresponding to high accuracy. More results are delayed to Appendix.}
  \label{matrix}
\end{figure*}

Methods are proposed to improve the robustness of DNNs against adversarial attacks, such as model pruning~\cite{madaan2020adversarial}, model quantization~\cite{lin2019defensive}, feature denoising~\cite{xie2019feature}, input purification~\cite{naseer2020self}, and adversarial training~\cite{madry2017towards}. 
Among them, adversarial training is one of the most effective approaches to defend against adversarial attacks. 
Adversarial training is a data augmentation technique that directly inputs the adversarial examples to each training step rather than natural examples. 
Adversarial examples are crafted by gradient descent algorithm and are bounded by $L_p$-norm around natural examples. 
The explicit purpose of adversarial training is to enable the DNN to obtain consistent latent feature representations in the face of adversarial and natural examples.

To better understand latent features relation under adversarial setting, we first empirically analyzed the latent feature of both adversarial and natural examples, and further found the similarity matrix of natural examples is more compact than those of adversarial examples.
\Cref{matrix} illustrates the latent features cosine similarity matrices on the first batches of CIFAR10 test dataset (batch-size is 256). 
From \Cref{matrix}, we can clearly observe that the block-wise pattern on the diagonal is more distinctive for natural examples, reflecting compact latent features relation. 

Motivated by this empirical observation, we propose latent feature relation consistency (LFRC) to improve the robustness of neural networks. 
Our framework is shown in \Cref{framework}. 
Specifically, given a mini-batch of natural examples drawn from the natural data space, we input the data into a neural network and obtain their latent features. 
Then the latent features are sequentially passed through a global average pooling (GAP) layer and a normalization layer to obtain $L_2$-normed representation. 
Finally, we can get the cosine similarity matrix of these latent features in this mini-batch. 
Following the same process, the adversarial similarity matrix can also be obtained by inputting adversarial examples (crafted by Project Gradient Descent (PGD) attack\cite{madry2017towards}) into the neural network. 
Penalizing the difference between the two similarity matrices makes the model learn consistent latent feature representation, thus improving the robustness of the model. 
It is worth mentioning that our proposed regularization technique is orthogonal to previous adversarial training methods.
The robustness of existing adversarial training methods can be improved to a certain extent by combining them with LFRC. 
Our contributions are summarized as follows:

\begin{itemize}
    \item We empirically analyze the latent feature relation of both adversarial and natural examples and find that the similarity matrix of natural examples is more compact than those of adversarial examples.
    
    \item We propose a novel training regularization technique: latent feature relation consistency (LFRC) which constrains the relation of adversarial examples in latent space to be consistent with the natural examples to improve the robustness of DNNs.

    \item We conduct extensive experiments on popular benchmark datasets and evaluate the performance against state-of-the-art attacks, which demonstrates our method can improve the robustness of neural networks by a large margin.
\end{itemize}

%-------------------------------------------------------------------------
\section{Related Work}
\subsection{Adversarial attack}
Since~\cite{goodfellow2014explaining} finds the adversarial perturbations in neural networks, lots of attack methods have been proposed to craft adversarial examples which can fool deep neural networks. 
Based on the accessibility to the knowledge of the model, attacks can be divided into white-box and black-box attacks. 
The white-box attack generated adversarial examples with the knowledge of the target model, while the black-box attack is agnostic to the model information.
The most famous white-box attack is Fast Gradient Sign Method (FGSM)~\cite{goodfellow2014explaining}, which only needs one step to create adversarial examples. 
Further,~\cite{madry2017towards} proposed the Project Gradient Descent (PGD) attack, which is the most used adversarial attack in the adversarial research field. 
Black-box attacks can be further divided into transfer-based attacks and query-based attacks. 
Transfer-based attacks are performed by generating adversarial examples on a surrogate model and then transferring them to the target model. 
The main work in this area is to enhance the transferability of adversarial examples, \eg, momentum attack\cite{dong2018boosting}, Scale-invariance~\cite{lin2019nesterov}, and Translation-invariance~\cite{dong2019evading} adversarial attack.
Query-based attacks are performed using examples that constantly query the hard or soft labels returned by the model, like SPSA~\cite{uesato2018adversarial} and Square~\cite{andriushchenko2020square}. 
AutoAttack (AA) is an ensemble attack, which consists of three white-box attacks (APGD-CE~\cite{croce2020reliable}, APGD-DLR~\cite{croce2020reliable}, and FAB~\cite{croce2020minimally}) and one black-box attack (Square attack~\cite{andriushchenko2020square}) for a total of four attack methods. 
AA is a parameter-free method that does not require specifying any parameters and is undoubtedly the most powerful attack method to date.

\subsection{Adversarial robustness}
Along with the development of adversarial attacks, there is also a lot of work to improve the adversarial robustness of deep neural networks~\cite{he2016deep}. 
Adversarial training~\cite{madry2017towards} is the most effective way to defend against adversarial examples. 
A body of work uses new objective functions and regularizations to further improve the effectiveness of standard adversarial training.
TRADES~\cite{zhang2019theoretically} analysis the trade-off between clean accuracy and adversarial accuracy, and uses Kullback-Leibler divergence~\cite{kullback1951information} to balance them. 
Misclassification Aware adveRsarial Training (MART)~\cite{wang2019improving} emphasizes the misclassified examples. 
Adversarial logit pairing (ALP)~\cite{kannan2018adversarial} defines a pairing loss that pulls adversarial logit and natural logit together.
Adversarial Neural Pruning (ANP)~\cite{madaan2020adversarial} uses Bayesian methods to prune vulnerable features.
Feature Denoising (FD)~\cite{xie2019feature} proposes a block to purify the adversarial feature map and obtain state-of-the-art on ImageNet. 
\cite{zhang2020geometry} uses sample reweighting techniques to improve the adversarial robustness against PGD attack, but not against AutoAttack.

\section{Method}
In this section, we first introduce the Standard Adversarial Training (SAT) proposed by Madry \etal~\cite{madry2017towards}. 
Then, we will describe the Latent Feature Relation Consistency (LFRC) in detail.
Finally, we will introduce the training procedure.
Our framework is shown in \Cref{framework}.

\subsection{Standard Adversarial Training}
Given a standard training data set $D = {(x_i,y_i)_{i=1}^n}$ with n examples and k classes, where $x_i \in \mathbb{R}^d$ is the natural example and $y_i=\{1,2,\dots,k\}$ is corresponding ground-truth label. 
Standard training achieves good classification performance by minimizing the empirical risk on the training dataset, formulated as \cref{st}:
\begin{equation}
    \underset{\theta \in \Theta}{\min} \  \frac{1}{n}\sum_{i=1}^{n} \mathcal{L}(f_{\theta}(x_i) , y_i),
    \label{st}
\end{equation}
where $\mathcal{L}$ is the cross-entropy loss widely used in classification tasks, $f:\mathbb{R}^d \rightarrow \mathbb{R}^k$ is the neural network parameterized by $\theta$ and $\Theta$ is the parameter space of $\theta$. 
However, the neural network trained in this way does not have good prediction accuracy for adversarial examples. 
To solve this problem, Madry \etal~\cite{madry2017towards} used adversarial data to train the neural network, which can be formulated as a min-max optimization problem:
\begin{equation}
    \underset{\theta \in \Theta}{\min} \ \underset{x'_i \in D'}{\max} \ \frac{1}{n}\sum_{i=1}^{n} \mathcal{L}(f_{\theta}(x'_i), y_i).
    \label{sat}
\end{equation}

$D'$ is the sampling space of the adversarial example $x'_i$, which is bounded in the $L_p$-norm neighbor space of the natural example $x_i$, i.e., $D' = \{x'_i \ | \  ||(x'_i - x_i)||_p \leq \epsilon\}$. 
In this paper, only $p=\infty$ is considered.
Madry \etal~\cite{madry2017towards} used projective gradient descent for the solution of the internal maximization equation, formulated as \cref{pgd}:
\begin{equation}
    x'_{t+1} = \mathcal{P}(x'_t + \alpha \cdot \nabla_{x'_t} \mathcal{L}(f_{\theta}(x'_t), y)),
    \label{pgd}
\end{equation}
where $\mathcal{P}$ is the projection function that projects the adversarial data $x'_i$ into the norm space of natural example $x_i$. 
$\alpha$ is the step size and $t$ is the current number of iterations.

\subsection{Latent Feature Relation Consistency}
\label{LFRC}
Recall from \cref{matrix}, the similarity matrix of natural examples is more compact than adversarial examples, i.e., the similarity matrix of natural examples has brighter blocks on the diagonal than the adversarial similarity matrix, which indicates that the latent representations of natural examples from the same class are more concentrated than those of adversarial examples. 
Motivated by this empirical discovery, we want the relation of adversarial features to be as good as natural features', thus achieving better feature relations for classification.
The correlation between feature matrices difference and accuracy difference is studied in \cref{correlation}. 
Specifically, we penalize the difference between these two similarity matrices to enable the model to learn consistent latent feature representations, thus improving the robustness of the model. 
LFRC sheds light on the latent feature relations under an adversarial setting which was neglected by previous methods.

Specifically, given a mini-batch natural example $X$, the latent features at layer $l$ ($l \in \{1,2,\dots, L\}$) of the neural network $f_\theta$ are denoted as $f_\theta^l(X) \in \mathbb{R}^{B \times C \times H \times W}$, where $B$ is the input batch size, $C$ is the number of output channels, and $H$ and $W$ are the spatial dimensions. 
Similarly, the latent features of the adversarial example $X'$ (crafted by PGD on natural examples $X$) at the corresponding $l$-th layer are denoted as $f_\theta^l(X')$. 
For computational efficiency, we first perform global average pooling on the latent features $f_{\theta}^l(X)$ to obtain the representations vectors $A^l(X) \in \mathbb{R}^{B \times C}$. 
Equationally, for the $c$-th channel:
\begin{equation}
    A_{b,c}^l(X) = \frac{1}{H \times W} \sum_{h=1}^{H} \sum_{w=1}^{W} f^l_{\theta}(X)_{[b,h,w,c]}.
    \label{gap}
\end{equation}

For the stability of the network's training and convenience of calculating similarity, $L_{2}$ normalization is applied on the channel dimension of $A^l(X)$ to obtain normalized representations $Q^l(X) \in \mathbb{R}^{B \times C}$:
\begin{equation}
    Q^l_{b,c}(X) = \frac{A^l_{b,c}(X)} {\sqrt{\sum_{c=1}^{C}  [A_{b,c}^{l}(X)] ^ 2}}.
\end{equation}
We can then calculate the feature similarity matrix of natural examples at the $l$-th layer:
\begin{equation}
    M^{l}(X) = Q^{l}(X) \cdot Q^{l}(X)^T,
\end{equation}
where $M^{l}(X) \in \mathbb{R}^{B \times B}$ is a symmetric matrix. 
$M_{i,j}$ denotes the cosine similarity of $X_i$ and $X_j$.
According to the same process, we feed-forward adversarial examples into the neural network and can also obtain the feature similarity matrix of the adversarial examples at the $l$-th layer:
\begin{equation}
    M^{l}(X') = Q^{l}(X') \cdot Q^{l}(X')^{T}.
\end{equation}

Then we penalize the difference between the two similarity matrices. The consistency loss at $l$-th layer is formulated as :
\begin{equation}
    \mathcal{L}^{l}_{LFRC}(X, X') = \frac{1}{B^2} \sum_{i=1}^{B} \sum_{j=1}^{B} e^{|M^{l}_{i,j}(X') - M^{l}_{i,j}(X)|}.
    \label{lossrc}
\end{equation}
Combining latent feature relation consistency loss with standard adversarial training strategy, the outer minimization formulation can be formulated as:
\begin{equation}
    \mathcal{L}_{total} = \mathcal{L}_{CE}(f_{\theta}(X') , Y) + \lambda \sum_{l=1}^{L} \ \mathcal{L}^{l}_{LFRC}(X,X'),
    \label{loss_total}
\end{equation}
where $\lambda$ is the hyperparameter balancing the normal loss and the latent feature relation consistency loss.

\begin{figure}[t]
    \begin{center}
    \includegraphics[width=\columnwidth]{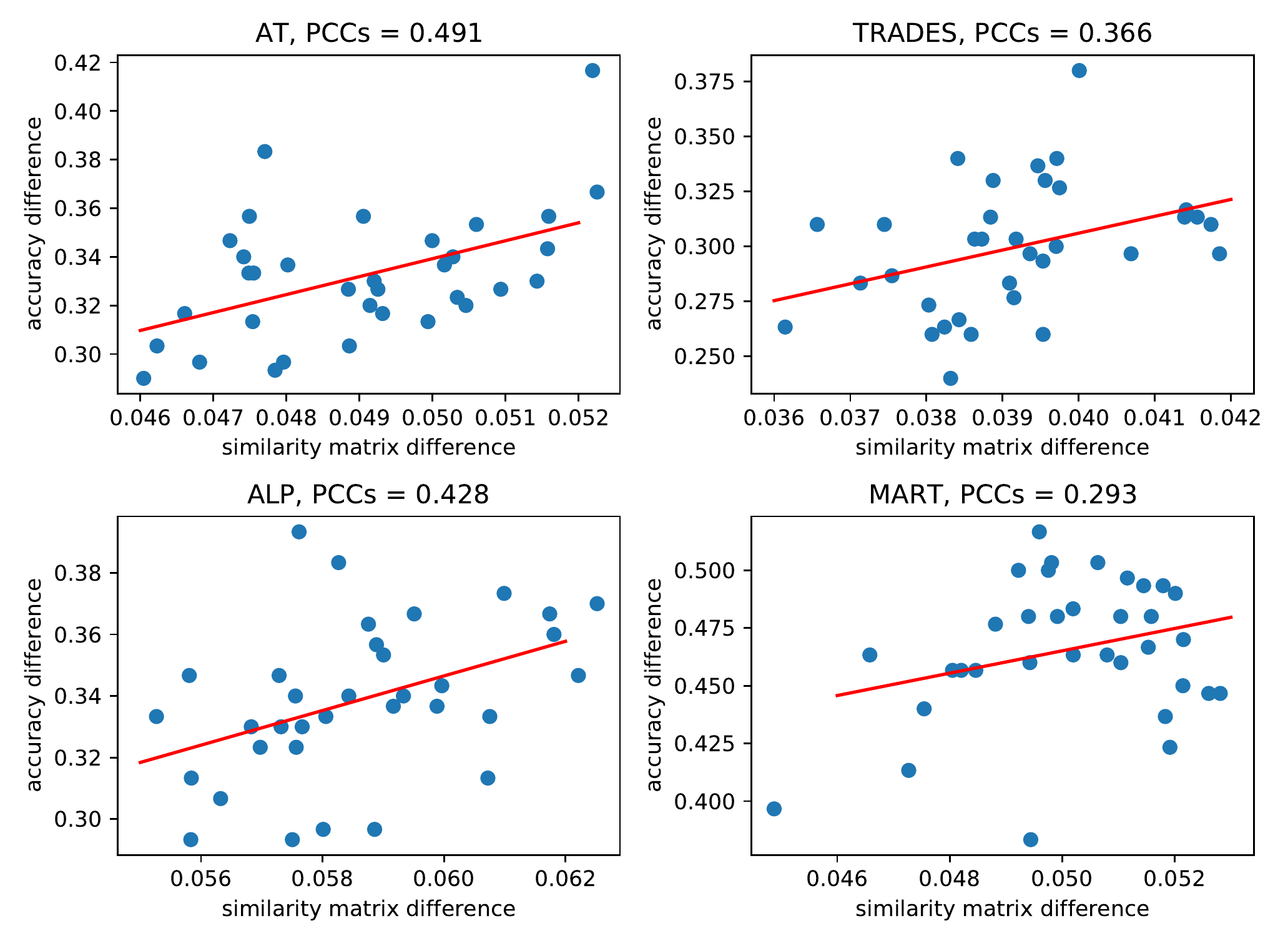}
    \end{center}
    \caption{Correlation of similarity matrix difference and accuracy difference between adversarial and natural examples obtained by different methods (AT~\cite{madry2017towards}, TRADES~\cite{zhang2019theoretically}, ALP~\cite{kannan2018adversarial}, MART~\cite{wang2019improving}). The red line is obtained by the least squares method. At the top of each plot, we mark the Pearson correlation coefficients (PCCs) of similarity matrix difference and accuracy difference.}
    \label{curve}
\end{figure}

\subsection{Correlation}
\label{correlation}
In \Cref{curve}, we plot the relationship between the latent feature similarity matrix difference ($DS$) and the corresponding accuracy difference ($DA$) for the adversarial and natural examples on the entire CIFAR10 dataset. 
We use least squares to fit these points to obtain the red curve. 
Specifically, given a mini-batch natural example $X$ and the adversarial examples $X'$, the similarity matrix difference (horizontal axis) is calculated as:
\begin{equation}
    DS = \frac{1}{B^2} \sum_{i=1}^{B}\sum_{j=1}^{B} |M^{l}_{i,j}(X') - M^{l}_{i,j}(X)|,
\end{equation}
where $B$ is the batch size. For the same mini-batch, we can get the accuracy difference (vertical axis):
\begin{equation}
    DA = \sum_{i=1}^{B} \mathbb{I} (f(x_i),y_i) - \sum_{i=1}^{B} \mathbb{I}(f(x'_i),y_i),
\end{equation}
where $\mathbb{I}$ is an indicator function, which is 1 when $f(x_i)=y_i$ and 0 otherwise.

In addition, we calculated the Pearson correlation coefficients (PCCs) of the two variables:
\begin{equation}
    r = \frac{\sum_{i=1}^{n}(DS_i - \overline{DS})(DA_i - \overline{DA})}{\sqrt{\sum_{i=1}^{n}(DS_i - \overline{DS})^2}\sqrt{\sum_{i=1}^{n}(DA_i - \overline{DA})^2}}.
\end{equation}

The Pearson correlation coefficients (PCCs) for AT, TRADES, and ALP are 0.491, 0.366, and 0.428, respectively. 
A moderately strong correlation between accuracy differences and similarity matrix differences is found. 
The smaller difference in the similarity matrices, the smaller difference in accuracy between the adversarial and natural examples, and vice versa.

%-----------------------------algorithm-----------------------------------
\begin{algorithm}[t]
	\renewcommand{\algorithmicrequire}{\textbf{Input:}}
	\renewcommand{\algorithmicensure}{\textbf{Output:}}
	\caption{Adversarial training with Latent Feature Relation Consistency (LFRC)}
	\label{alg:1}
	\begin{algorithmic}[1]
		\REQUIRE Neural network $f_\theta$, input data space $D$, training epochs $T$, the number of batches N, batch-size m, loss balance hyperparameter $\lambda$, learning rate $\eta$.
		\ENSURE robust Neural network $f_\theta$.
		\FOR{epoch $\leq$ T}
		\FOR{batch $\leq$ N}
		\STATE Sample a mini-batch examples from input data space: $\{(x_i, y_i)\}_{i=1}^{m}$.
		\STATE Generate $x'$ by \cref{pgd}.
		\STATE Calculate empirical risk by \cref{loss_total}.
		\STATE $\theta \leftarrow \theta - \eta \nabla_{\theta} {\mathcal{L}}_{total}(f_{\theta}(x'), y)$.
		\ENDFOR
		\ENDFOR
		\STATE \textbf{return} robust model $f_{\theta}$
	\end{algorithmic}
\end{algorithm}
%-------------------------------------------------------------------------

\subsection{Training and Optimization}
The process of our algorithm is summarized in \cref{alg:1}. 
LFRC first generates adversarial examples based on natural examples by PGD attack, which is the same as \cref{pgd}. 
Then, we feed forward adversarial and natural examples in parallel into the network and calculate the cross-entropy loss and the latent feature relation consistency loss. 
Finally, the update of the neural network parameters is performed according to the stochastic gradient descent method. 
Repeating the above steps until training is over, the algorithm will output a robust neural network.

\subsection{Difference from previous methods}
Previous methods focused on the design of the loss, and few of them focused on the latent features. 
TRADES and ALP consider the pair (natural and adversarial) information as supervision, together with the true label.
In contrast, LFRC resorts to stronger guidance, \ie, latent feature structure of batch examples, by considering the relationship between features in a batch.
Our method constrains the difference between their similarity matrices so that the similarity relationship between adversarial examples is consistent with that between natural examples.
Overall LFRC is a simple but efficient regularization technique to improve the robustness of the model, which can be inserted in any position of the network.

\section{Experiment}
\label{experiment}
In this section, we conduct extensive experiments on popular benchmark datasets and networks to demonstrate the effectiveness of LFRC. 
First, we present the basic setup of the experiments in \Cref{setup} and report the white-box and black-box robustness respectively. 
LFRC is implemented at the last residual block of the ResNet and WideResNet networks by default, and \Cref{layer} has the ablation experiments on the constrained positions.

\subsection{Experimental Setup}
\label{setup}

\textbf{Baseline methods:}
LFRC improves adversarial robustness by constraining the latent feature relation between natural and adversarial examples, which can be combined with previous adversarial training strategies.
We specifically consider three adversarial training methods, which are widely used in adversarial training. 
The first is the standard adversarial training (AT) proposed by~\cite{madry2017towards}, which defines adversarial training as a min-max problem to find saddle points. 
The second is TRADES proposed by~\cite{zhang2019theoretically}, which uses a hyperparameter to balance natural and adversarial accuracy, and the method is the current sota. 
The third is adversarial logit pairing (ALP) proposed by~\cite{kannan2018adversarial}, which encourages a pair of examples with more similar logit. 
Implementation-wise, we strictly follow the parameter settings in the original paper.

\textbf{Training setup:}
\label{train_set}
Our overall training parameters refer to~\cite{madaan2020adversarial}. 
Specifically, we use SGD (momentum 0.9, batch size 128) to train ResNet18 for 100 epochs on the CIFAR10 dataset with weight decay 2e-4 and initial learning rate 0.1, divided by 10 at 75-th and 90-th epoch, respectively.
RandomCrop and RandomHorizontalFlip are used as data augmentation. 
For the internal maximization process, we use PGD$_{10}$ adversarial attack to solve, with a random start, step size 2.0/255, and perturbation size 8.0/255. 
We use $\lambda = 100$ in all experiments. 
The experimental parameters of ResNet18 in CIFAR100, WideResNet28-10 in CIFAR10, and CIFAR100 are the same as described above.

%------------------white ResNet18-CIFAR10-----------

\begin{table*}[ht]
    \begin{center}
    \begin{tabular}{c|c|ccccc|ccccc}
        \toprule
        \multirow{2}{*}{Dataset}&
        \multirow{2}{*}{Method} & \multicolumn{5}{c|}{\textbf{Best Checkpoint}}        & \multicolumn{5}{c}{\textbf{Last Checkpoint}}         \\
        && Clean & FGSM & PGD$_{20}$ & CW$_\infty$ & AA & Clean & FGSM & PGD$_{20}$ & CW$_\infty$ & AA \\ \midrule
        \multirow{6}{*}{CIFAR-10}
        &AT & \textbf{83.30} &\textbf{ 56.67} & 49.99 & 49.07 & 46.77 & \textbf{84.67}& 55.65 & 46.48 & 46.83 & 44.11   \\
        &\textbf{+LFRC}& 82.44 & 56.57 & \textbf{50.99} & \textbf{49.83} & \textbf{47.55} & 84.29 & \textbf{56.52} & \textbf{48.42} & \textbf{48.27} &\textbf{45.40}    \\ \cmidrule{2-12}
        &TRADES                   & 81.86 & 57.19 & 51.99 & 49.96 &48.30 & 82.36 & 57.55 & 51.51 & 49.87& 48.24    \\
        &\textbf{+LFRC}              & \textbf{82.43} & \textbf{58.21} & \textbf{52.82} & \textbf{50.98} &\textbf{49.39} & \textbf{82.64} & \textbf{58.15} & \textbf{52.17} & \textbf{50.29}&\textbf{48.74}    \\ \cmidrule{2-12}
        &ALP                     &\textbf{83.95}&55.95&50.15&48.61&47.42&84.59&56.47&48.03&47.21&45.32   \\
        &\textbf{+LFRC}       &83.33&\textbf{56.66}&\textbf{51.14}&\textbf{49.38}&\textbf{48.18}&\textbf{84.93}&\textbf{57.37}&\textbf{49.47}&\textbf{48.52}&\textbf{46.72}   \\
        \bottomrule
        \toprule
        \multirow{6}{*}{CIFAR-100}
        &AT &\textbf{57.05}&28.90&24.62&24.10&22.39&56.51&26.58&22.08&22.61&20.53    \\
        &\textbf{+LFRC}                 &55.71&\textbf{30.49}&\textbf{27.05}&\textbf{25.89}&\textbf{23.65}&\textbf{57.04}&\textbf{28.11}&\textbf{23.69}&\textbf{23.49}&\textbf{21.70}   \\ \cmidrule{2-12}
        &TRADES                  & 54.02&30.21 & 27.40 & 24.03 & 23.10 & 53.83& 29.99 & 27.14 &23.90&22.96    \\
        &\textbf{+LFRC}              &\textbf{55.85}&\textbf{31.21}&\textbf{28.66}&\textbf{25.47}&\textbf{24.55}&\textbf{55.52}&\textbf{31.00}&\textbf{28.41}&\textbf{25.14}&\textbf{24.18}    \\ \cmidrule{2-12}
        &ALP                     &58.57&27.97&23.61&22.76&21.10&58.19&27.76&23.23&22.34&20.79    \\
        &\textbf{+LFRC}                &\textbf{59.15}&\textbf{29.42}&\textbf{25.44}&\textbf{24.30}&\textbf{22.84}&\textbf{58.80}&\textbf{29.38}&\textbf{24.92}&\textbf{23.88}&\textbf{22.37}    \\
        \bottomrule
    \end{tabular}
    \end{center}
    \caption{The white-box robustness results(accuracy $(\%)$) of ResNet18 on CIFAR10 and CIFAR100, we report the results of the best checkpoint and last checkpoint following previous work~\cite{madry2017towards}. The best results are marked using \textbf{boldface}. ``\textbf{+LFRC}" denotes the combination of our method with the previous method}
    \label{white-ResNet}
\vspace{-0.2cm}
\end{table*}

\textbf{Evaluation setup:}
\label{eval_set}
We report the clean accuracy on natural examples and the adversarial accuracy on adversarial examples. 
For adversarial accuracy, we report both white-box and black-box. 
We follow the widely used protocols in the adversarial research field. 
For the white-box attack, we consider three basic attack methods: FGSM~\cite{goodfellow2014explaining}, PGD~\cite{madry2017towards}, and CW$_{\infty}$~\cite{carlini2017towards} optimized by PGD$_{20}$, and a stronger ensemble attack method named AutoAttack (AA)~\cite{croce2020reliable}.
For the black-box attacks, we consider both transfer-based attacks and query-based attacks.

\subsection{White-box Robustness}
For FGSM, PGD, CW$_\infty$, and AA, the attack perturbation budget is all 8.0/255, and the step size for PGD, CW$_\infty$ is 2.0/255 with 20 iterations. 
Following the experimental report of the previous paper, we report the results both at the best checkpoint and the last checkpoint. 
The best checkpoint results are selected based on their defense performance on the CIFAR10 test set for PGD (attack step size 2.0/255, perturbation budget 8.0/255, and iteration number 10).

\cref{white-ResNet} report the white-box test results of ResNet18 on CIFAR10 and CIFAR100, respectively. 
Experimental results show that our method combined with the previous method improves the robustness of the model for all four attacks on both datasets, which demonstrates the generality of our method. 
For example, for the robustness of PGD$_{20}$,
``AT+LFRC" improves the robustness by 1.0\% on CIFAR10 and 2.43\% on CIFAR100 respectively compared to AT. 
``TRADES+LFRC" improves the robustness by 0.9\% on CIFAR10 and 1.26\% on CIFAR100 respectively compared to TRADES. 
The improvement effect is more obvious on AutoAttack, which is the state-of-the-art of current adversarial attack method. 

Moreover, we observe two phenomena. First, LFRC can improve the prediction accuracy of natural examples in some cases, which is uncommon in the previous adversarial training methods.
Second, compared with CIFAR10, the improvement of LFRC on CIFAR100 is more stable and significant. 
Basically, it can improve the accuracy under all prediction conditions and the improvement effect is more significant.

%-----------------------------------------blackResNet18-CIFAR10-----------------------------
\begin{table*}[ht]
    \begin{center}
    \begin{tabular}{c|c|ccccc|ccccc}
        \toprule
        \multirow{2}{*}{Dataset}&
        \multirow{2}{*}{Method} & \multicolumn{5}{c|}{\textbf{Best Checkpoint}}   & \multicolumn{5}{c}{\textbf{Last Checkpoint}}    \\
                                && FGSM & PGD$_{20}$ & PGD$_{40}$ & CW$_\infty$&Square & FGSM & PGD$_{20}$ & PGD$_{40}$ & CW$_\infty$ & Square \\ \midrule
        
       \multirow{6}{*}{CIFAR-10}& AT                      &\textbf{64.45}&61.23&61.22&60.77&54.37 & 65.34&61.92&61.87&61.55&53.06     \\
        &\textbf{+LFRC}                 &64.12&\textbf{61.62}&\textbf{61.63}&\textbf{61.29}&\textbf{54.55} &\textbf{65.48}&\textbf{62.47}&\textbf{62.39}&\textbf{62.21}&\textbf{53.40}    \\ \cmidrule{2-12}
        &TRADES                   &64.25&62.15&62.08&61.88& 54.90 &64.91&62.72&\textbf{62.86}&62.42& 54.96    \\
        &\textbf{+LFRC}              & \textbf{64.58}&\textbf{62.54}&\textbf{62.69}&\textbf{62.12}& \textbf{55.65} &\textbf{65.32}&\textbf{62.80}&62.81&\textbf{63.01}&\textbf{55.46}     \\ \cmidrule{2-12}
        &ALP                     &64.56&61.99&61.92&61.40&54.58&65.55&62.96&62.92&62.39&54.07   \\
        &\textbf{+LFRC}      &\textbf{65.16}&\textbf{62.91}&\textbf{62.65}&\textbf{62.69}&\textbf{55.09}&\textbf{66.19}&\textbf{63.37}&\textbf{63.27}&\textbf{63.32}&\textbf{54.91}    \\
        \bottomrule
        \toprule
        \multirow{6}{*}{CIFAR-100}
        &AT                      & 38.14&36.21&36.17&37.83& 27.18 &37.52&36.06&35.96&37.63&25.17      \\
        &\textbf{+LFRC}                 &\textbf{38.39}&\textbf{37.22}&\textbf{37.10}&\textbf{38.47}&\textbf{28.95} &\textbf{37.93}&\textbf{36.73}&\textbf{36.85}&\textbf{38.39}&\textbf{26.70}\\ \cmidrule{2-12}
        &TRADES                   &37.76&37.00&36.99&37.93& 27.13 &37.69&36.59&36.52&37.65& 27.00 \\
        &\textbf{+LFRC}              &\textbf{38.69}&\textbf{37.61}&\textbf{37.61}&\textbf{38.91}&\textbf{28.54} &\textbf{38.52}&\textbf{37.43}&\textbf{37.44}&\textbf{38.72}&\textbf{28.09}\\ \cmidrule{2-12}
        &ALP                     &38.21&36.76&36.78&37.99& 26.36&38.40&36.82&36.93&38.15&25.86     \\
        &\textbf{+LFRC}                &\textbf{39.08}&\textbf{37.79}&\textbf{37.74}&\textbf{38.78}&\textbf{28.11}&\textbf{39.48}&\textbf{38.17}&\textbf{38.13}&\textbf{39.01}&\textbf{27.32}     \\
        \bottomrule
    \end{tabular}
    \end{center}
    \caption{The black-box robustness results(accuracy $(\%)$) of ResNet18 on CIFAR10 and CIFAR100, we report the results of best checkpoint and last checkpoint following previous work~\cite{madry2017towards}. The best results are marked using \textbf{boldface}. ``\textbf{+LFRC}" denotes the combination of our method with the previous method.}
    \label{black-resnet}
\end{table*}

%-------------------------------------------------------

\subsection{Black-box Robustness}
We use both transfer-based attacks and query-based attacks to perform the black-box tests. 
For the transfer-based attack, we use the PGD$_{10}$ adversarially trained ResNet34 network as the surrogate model (the training parameters are the same as \Cref{train_set}). 
Once the surrogate model is trained, the adversarial examples are generated on the surrogate model using four adversarial attacks (FGSM, PGD$_{20}$, PGD$_{40}$ and CW$_\infty$), and then transferred to the target model for adversarial testing. 
For query-based attacks, the most used one is Square~\cite{andriushchenko2020square} which is a query-efficient attack.

\cref{black-resnet} show the black-box adversarial results of ResNet18 on CIFAR10 and CIFAR100 under different conditions. 
Our LFRC improves model robustness against black-box attacks consistently. 
For example, for the robustness of PGD$_{20}$ generated by ResNet34,
``AT+LFRC" improves the robustness by 0.39\% on CIFAR10 and 0.99\% on CIFAR100 respectively compared to AT. 
``TRADES+LFRC" improves the robustness by 0.39\% on CIFAR10 and 0.61\% on CIFAR100 respectively compared to TRADES.
Similarly, LFRC has a more pronounced boosting effect for CIFAR100 compared to CIFAR10, as can be seen from the query-based attack Square.

\section{Ablation Study}
\label{ablation}
In this section, we conduct extensive ablation experiments to get more insight into our method LFRC. 
Note: the ablation experiments are all trained on CIFAR10 using ResNet18 neural network. The other parameters remain the same as described in \Cref{train_set}.

\begin{table}[t]
    \begin{center}
    \setlength{\tabcolsep}{4.7mm}
    \begin{tabular}{c|c|cc}
        \toprule
        Method                   & Position & PGD$_{20}$ & AA \\
        \midrule
        \multirow{4}{*}{AT+LFRC} & Block1   & 49.40&45.91  \\ 
        & Block2   &50.24&46.52   \\ 
         & Block3   &50.71&46.36 \\
         & Block4   &\textbf{50.99}&\textbf{47.55}   \\
        \bottomrule
    \end{tabular}     
    \end{center}
    \caption{The robustness of LFRC inserted behind different residual blocks of ResNet18. We test on the best checkpoint.}
    \label{abla-layer}
\end{table}

\subsection{Different Layers}
\label{layer}
LFRC improves model robustness by constraining the similarity matrix of latent layer features. 
For neural networks containing many latent layers, constraining the similarity matrix of features at different layers may have different effects on robustness. 
To investigate the effect of LFRC constraint position on adversarial robustness, we inserted LFRC at different positions of the neural network. 
Specifically, for the ResNet18 network architecture, we inserted LFRC after four different residual blocks to explore the effect of LFRC inserted position on white-box robustness.

\cref{abla-layer} shows the experimental results of inserting the LFRC behind different residual blocks of ResNet18. 
It can be observed that LFRC inserted in the last block achieves the best results. 
We assume that placing LFRC at the last level can constrain latent feature similarity relationships efficiently. 
Constraining only the lower-level latent features does not necessarily guarantee that the higher-level latent features also maintain the corresponding similarity relationship, which can be seen from the experimental results. 
Moreover, inserting LFRC after block 2 and block 3 both can improve the adversarial accuracy against PGD$_{20}$.

\begin{table}[t]
    \begin{center}
    \setlength{\tabcolsep}{6.5mm}
    \begin{tabular}{c|c|c}
        \toprule
        Method      & PGD$_{20}$ & AA \\ \midrule
        AT          & 49.99&46.77   \\ 
        AT+AWP      & 54.02 & 49.12 \\ 
        AT+LFRC     & 50.99&47.55    \\
        AT+LFRC+AWP & \textbf{54.35}& \textbf{50.18}   \\
        \bottomrule
    \end{tabular}
    \end{center}
    \caption{The results (acc\%) of LFRC and AWP working together on ResNet18.
    We test on the best checkpoint.}
    \label{abla-awp}
\end{table}

\subsection{Orthogonality}
\label{awp}
Previous results demonstrate that LFRC is orthogonal to adversarial training methods of designing objective functions.
Adversarial Weight Perturbation(AWP)~\cite{wu2020adversarial}, can also be seen as orthogonal to the other methods, which applies perturbation to the weights of the neural network to narrow the generalizability gap of the model and thus enhance the robustness.
They experimentally explored the robustness results of different methods combined with AWP, and most of them were able to obtain improvements. 
In this section, we combine LFRC and AWP to explore their interactions.

Table \ref{abla-awp} shows the results of combining our method with AWP. 
We observe that combining AWP and LFRC can further improve the adversarial robustness of neural networks, which indicates that our method and AWP are orthogonal.

\begin{figure}
    \begin{center}
    \includegraphics[width=\columnwidth]{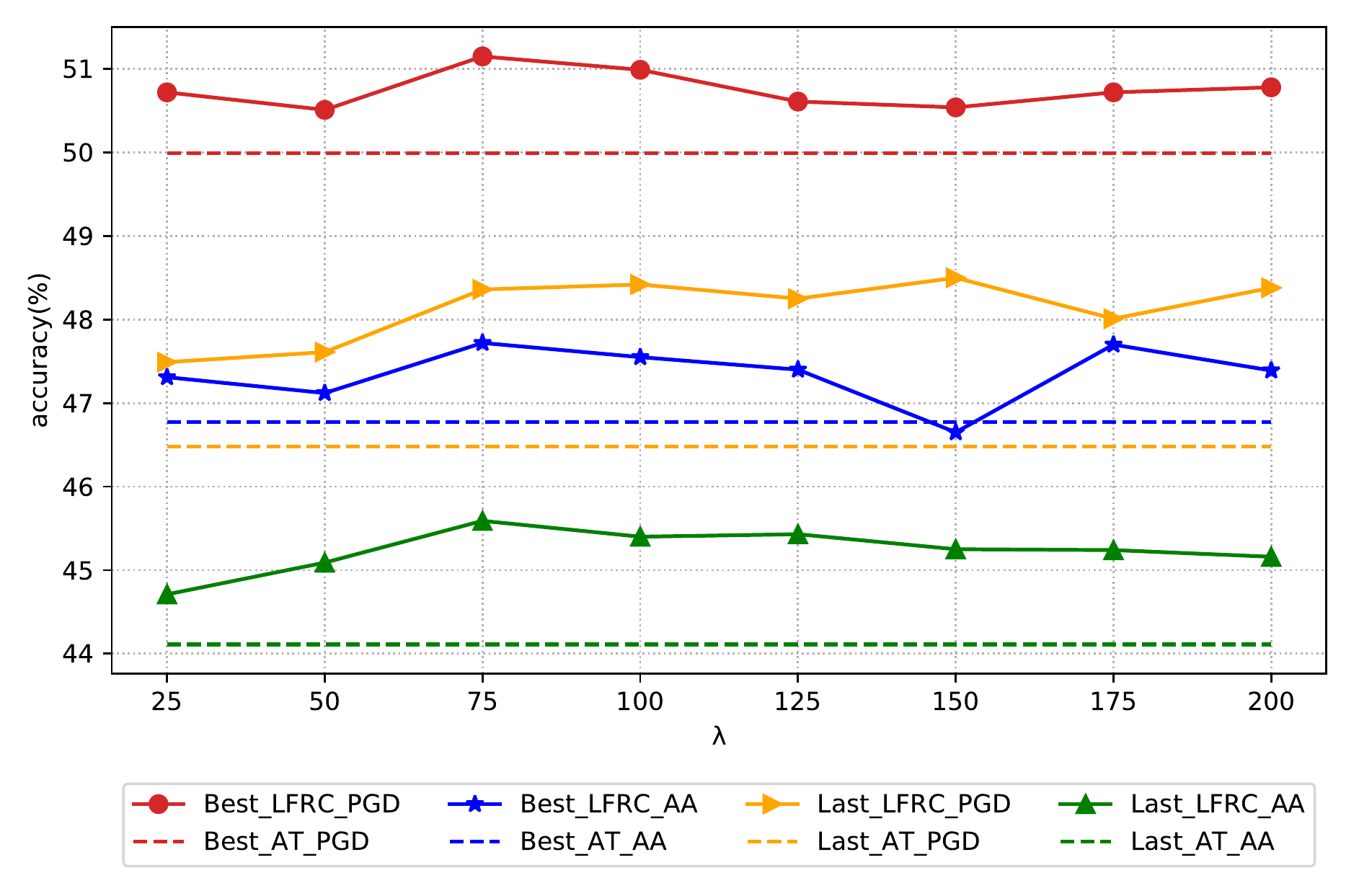}
    \end{center}
    \caption{The effect of hyperparameter $\lambda$ on white-box robustness. \textbf{``Best-LFRC-AA"} denotes the test of AutoAttack using the model with the best checkpoint trained with LFRC, and the other seven representations are the same.}
    \label{abla-lambda}
\end{figure}
\subsection{Hyperparameter}
\label{lambda}

As described in \Cref{loss_total}, $\lambda$ plays a balancing role between the normal and latent feature consistency loss. 
We explore the effect of the hyperparameter $\lambda$ on the adversarial robustness.
\Cref{abla-lambda} illustrates our results. 
Here we use several parameters varying from 0 to 200.
We report the white-box results of PGD$_{20}$ and AA for both the best checkpoint and the last checkpoint. 
Our method is not particularly sensitive to hyperparameter $\lambda$, and the robustness under all hyperparameters (shown by the solid line) is better than the baseline model (shown by the dashed line).

\begin{figure}
    \begin{center}
    \includegraphics[width=\columnwidth]{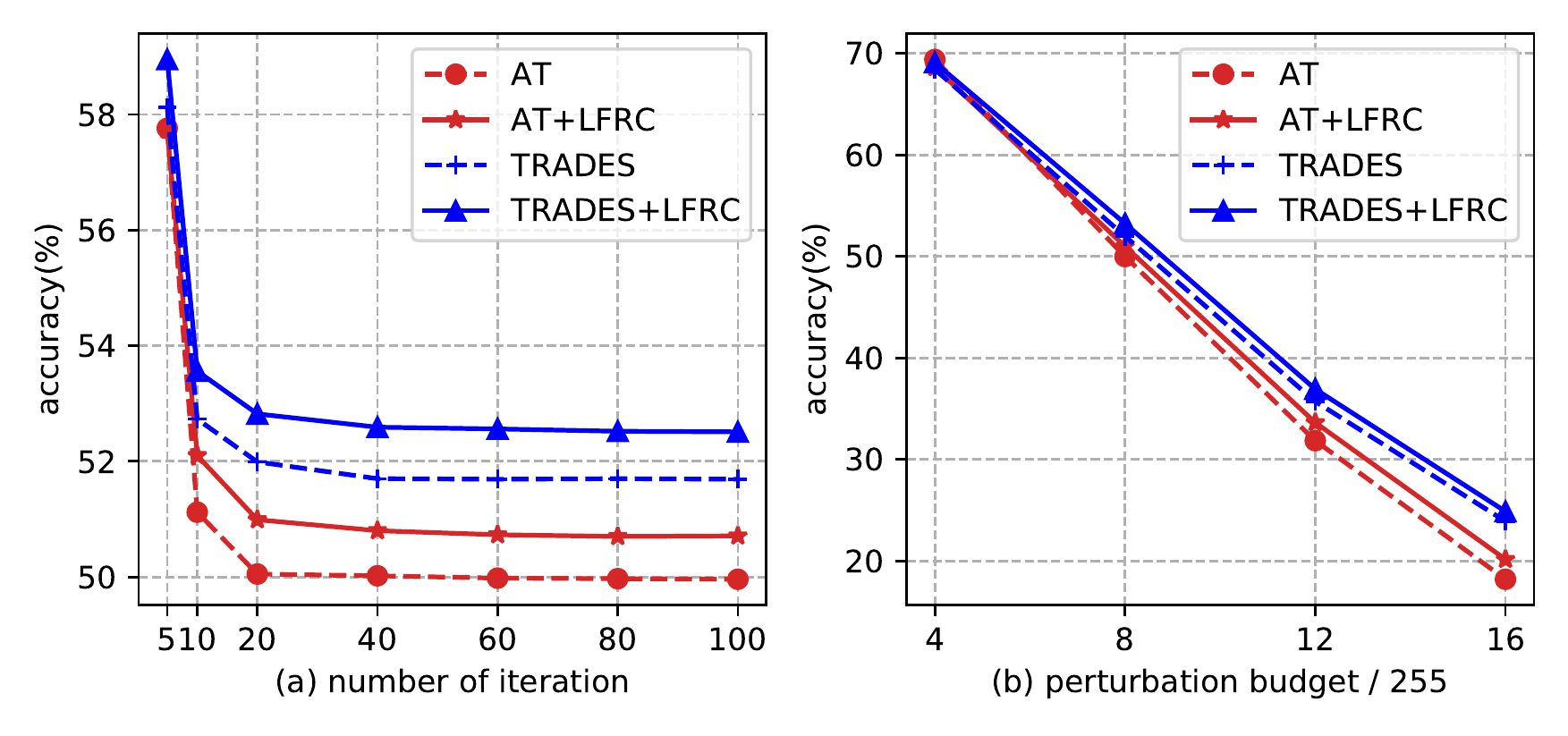}
    \end{center}
    \caption{(a) Robust accuracy(\%) of LFRC and the baselines on CIFAR10 under different iteration numbers. The step size is 2/255 and the perturbation budget is 8/255.
    (b) Robust accuracy(\%) of LFRC and baselines on CIFAR10 under different perturbation budgets. The step size is 2/255 and the iteration number is 20.}
    \label{abla-budget}
\end{figure}

\subsection{Different Budgets and Iterations}
\label{budget}
To demonstrate the effectiveness of LFRC, we conducted PGD experiments with different perturbation budgets and several iterations. 
The experimental results are illustrated in \Cref{abla-budget}. \Cref{abla-budget}(a) illustrates the robust accuracy(\%) of PGD attack for LFRC and baseline models under different numbers of iterations (varying from 5 to 100) with an attack budget of 8/255 and a step size of 2/255. 
\Cref{abla-budget}(b) illustrates the robust accuracy(\%) of PGD attack for LFRC and baseline models under different perturbation budgets (varying from 4/255 to 16/255) with an attack iteration number of 20 and a step size of 2/255. 
Our method LFRC can improve model robustness in various situations. 
Our method consistently improves robustness by 0.8\% under different perturbation budget settings.

%--------------------------------------------------------------------------
\section{Discussion and Future work}
\label{discussion}
Our approach works well on various datasets and networks by penalizing the differences between the similarity matrices of adversarial and natural examples. 
The idea comes from the fact that the similarity matrices of natural examples are more compact than adversarial examples, as illustrated in \Cref{matrix}. 
In the similarity matrices of \Cref{matrix}, we also observe an interesting phenomenon for the CIFAR10 dataset. 
There are ten classes $\{0,1,\dots,8,9\}$ corresponding to the ten small bright blocks on the diagonal (top left to bottom right) of the similarity matrix. 
However, the brightness of the blocks in different classes is different.
The blocks of classes $\{0,1,8,9\}$ are relatively brighter, representing more concentrated latent features in these classes. 
The presence of a large bright block in the middle of the similarity matrix (classes $\{2,3,4,5,6,7\}$) demonstrates that the latent features of these classes are easily confused.
In recent years, there has been researched on sample reweighting~\cite{gao2021local,zhang2020geometry} in the adversarial robustness field.
We hope this phenomenon brings new insights to sample reweighting, i.e., examples in easily confused classes should be given greater weight.

%-------------------------------------------------------------------------
\section{Conclusion}
\label{conclusion}
In this paper, we investigate the latent feature relationship between natural and adversarial examples.
We empirically found that the latent features similarity matrix of natural examples is more compact than those of adversarial examples for the adversarially trained model. 
Based on this observation, we creatively propose Latent Feature Relation Consistency (\textbf{LFRC}), a regularization technique that constrains the similarity matrices of adversarial and natural examples to improve the adversarial robustness of neural networks. 
When combined with the previous adversarial training method, the adversarial accuracy of both the white-box and black-box can be further improved. 
We have also made many insightful understandings of LFRC: inserted position and hyperparameter.
In summary, our method is a simple but efficient approach that can be easily combined with other methods to obtain further improvements.

%-------------------------------------------------------------------------

%-------------------------------------------------------------------------

%-------------------------------------------------------------------------

%------------------------------------------------------------------------

%%%%%%%%% REFERENCES
{\small
\bibliographystyle{ieee_fullname}
\bibliography{egbib}
}

%--------------------------appendix----------------

\clearpage
\newpage
\appendix

\renewcommand\thesection{\Alph{section}} 
\renewcommand\thesubsection{\Alph{section}.\arabic{subsection}} 
\renewcommand\thefigure{\Alph{section}\arabic{figure}} 
\renewcommand\thetable{\Alph{section}\arabic{table}} 
\setcounter{section}{0}
\setcounter{figure}{0}	
\setcounter{table}{0}

\section{Additional Experiments}

\subsection{Metric functions}
In our experiments, we use the exponential function as the metric function of the two similarity matrices. We also used $L_1$ norm (\Cref{lossl1}) and $L_2$ norm (\ref{lossl2}) metric functions for our experiments.
The results are shown in \cref{metric}.
As can be seen from the results, a relatively large improvement can be obtained by using exponential function as the metric function.

\begin{equation}
    \mathcal{L}^{l}_{LFRC}(x, x') = \frac{1}{B^2} \sum_{i=1}^{B} \sum_{j=1}^{B} {|M^{l}_{i,j}(x') - M^{l}_{i,j}(x)|}.
    \label{lossl1}
\end{equation}

\begin{equation}
    \mathcal{L}^{l}_{LFRC}(x, x') = \frac{1}{B^2} \sum_{i=1}^{B} \sum_{j=1}^{B} {|M^{l}_{i,j}(x') - M^{l}_{i,j}(x)|^2}.
    \label{lossl2}
\end{equation}

\subsection{Standard training + LFRC.}
We report the white-box robustness of standard training and its combination with LFRC in \cref{tab:standard}.
The improvement is more significant than the combination of LFRC with previous adversarial methods.
Combining LFRC with standard training makes the latent feature relation of adversarial examples consistent with the relation of natural examples, which is more beneficial to classification.

\subsection{Experiment results on VGG-16}
We provide adversarial robustness of VGG-16 on CIFAR-10 in \cref{tab:vgg}.
LFRC can improve the robustness of VGG-16 as well as ResNet, which valid the generalizability of our LFRC.

\subsection{Experiment results on WideResNet28-10}
We conducted similar experiments on WideResNet28-10~\cite{zagoruyko2016wide} using AT, AT+LFRC, TRADES, TRADES+LFRC, ALP, ALP+LFRC. 
The training parameters were set the same as those used for ResNet18~\cite{he2016deep}. 
We inserted LFRC in the last residual block and used $\lambda = 100$ in all experiments. 
We report the experimental results at the best checkpoint and the last checkpoint.
\cref{white-ResNet-CIFAR10,white-ResNet-CIFAR100} show the results of WideResNet28-10 on CIFAR10 and CIFAR100, respectively. 
In most cases, LFRC can improve the robustness of the baseline model, further demonstrating the effectiveness of our proposed method.
Moreover, similar to ResNet18, the LFRC has a more pronounced enhancement effect on CIFAR100 than CIFAR10.

\begin{table}[t]
    \begin{center}
    \setlength{\tabcolsep}{3.5mm}
    \begin{tabular}{c|cc|cc}
        \toprule
        \multirow{2}{*}{Metric} & \multicolumn{2}{c|}{\textbf{Best Checkpoint}} & \multicolumn{2}{c}{\textbf{Last Checkpoint}} \\
        & PGD$_{20}$ & AA &PGD$_{20}$&AA \\
        \midrule
        $L_1$      & 50.25&46.63&47.74&44.64 \\
        $L_2$     &50.43&47.01&46.16&43.92    \\
         exp          &\textbf{50.99}&\textbf{47.56}&\textbf{48.42}&\textbf{45.40} \\
        \bottomrule
    \end{tabular}
    \end{center}
    \caption{The results of LFRC using different metric functions.
    We test on the best checkpoint and the last checkpoint.}
    \label{metric}
\end{table}

\begin{table}[t]
    \begin{center}
    \begin{tabular}{c|c|c|c}
        Method & FGSM & PGD & CW$_\infty$\\
        \midrule
        Standard & 19.26 &0.00&0.00  \\
        +LFRC& \textbf{56.72} & \textbf{50.12} & \textbf{49.98}
    \end{tabular}
    \end{center}
    \caption{White-box robustness of ResNet-18 on CIFAR-10.}
    \label{tab:standard}
\end{table}

\begin{table}[t]
    \begin{center}
    \begin{tabular}{c|c|c|c}
        Method & FGSM & PGD & CW$_\infty$\\
        \midrule
        AT & 51.87&42.43&43.63  \\
        +LFRC& \textbf{52.65} & \textbf{43.39} & \textbf{44.56}
    \end{tabular}
    \end{center}
    \caption{White-box robustness of VGG-16 on CIFAR-10.}
    \label{tab:vgg}
\end{table}

\subsection{Comparison with contrastive adversarial training.}
We discuss the related adversarial training works based on contrastive learning as follows.
SLAT~\cite{slat} is a single-step adversarial training method, which leverages the gradients of latent representation as adversarial perturbation.
LID~\cite{lid} tackles adversarial challenges by characterizing the dimensional properties of adversarial regions, via the use of Local Intrinsic Dimensionality.
LID is a discriminative  method, which is another widely researched field, which is totally different from adversarial training.
Recently, researchers use contrastive learning~\cite{rocl,acl} work to improve adversarial robustness.
These studies work in a way that pulls the positive pair and pushes the negative pair.
While LFRC improves robustness by making the relations of adversarial examples in a mini-batch consistent with the relations of natural examples in the same mini-batch.
Besides, LFRC is more powerful than these works.

\begin{table}[h]
    \begin{center}
    \begin{tabular}{ccc}
        RoCL~\cite{rocl} & ACL~\cite{acl} & LFRC\\
        \hline
        49.66 &49.36 & 50.99
    \end{tabular}
    \end{center}
\end{table}

%----------------------------------white ResNet18-CIFAR10-----------------------------
\begin{table*}[ht]
    \begin{center}
        \begin{tabular}{c|ccccc|ccccc}
        \toprule
        \multirow{2}{*}{Method} & \multicolumn{5}{c|}{\textbf{Best Checkpoint}}        & \multicolumn{5}{c}{\textbf{Last Checkpoint}}         \\
                                & Clean & FGSM & PGD$_{20}$ & CW$_\infty$&AA& Clean & FGSM & PGD$_{20}$ & CW$_\infty$ & AA \\ 
                                \noalign{\smallskip}\hline\noalign{\smallskip}
        AT                      &86.31&58.63&50.89&51.25&48.87&\textbf{86.19}&56.22&46.28&46.75&44.81   \\
        AT\textbf{+LFRC}        &\textbf{85.46}&\textbf{59.27}&\textbf{51.52}&\textbf{51.64}&\textbf{49.46}&85.16&\textbf{58.10}&\textbf{47.51}&\textbf{47.97}&\textbf{45.56} \\ 
        \noalign{\smallskip}\hline\noalign{\smallskip}
        TRADES                   &83.98&59.70&54.33&52.96&51.60&85.09&59.38&50.46&50.92&48.64    \\
        TRADES\textbf{+LFRC}    &\textbf{84.45}&\textbf{60.33}&\textbf{54.40}&\textbf{53.15}&\textbf{51.75}&\textbf{85.20}&\textbf{60.31}&\textbf{50.99}&\textbf{51.41}&\textbf{48.98}   \\ 
        \noalign{\smallskip}\hline\noalign{\smallskip}
        ALP                     &\textbf{85.75}&57.10&49.94&49.77&48.17&85.54&56.89&47.02&47.24&45.56   \\
        ALP\textbf{+LFRC}        &85.65&\textbf{57.31}&\textbf{50.93}&\textbf{50.21}&\textbf{48.77}&\textbf{85.84}&\textbf{57.58}&\textbf{47.93}&\textbf{48.15}&\textbf{46.47}       \\
        \bottomrule
    \end{tabular}
    \end{center}
    \caption{The white-box robustness results(accuracy $(\%)$) of WideResNet28-10 on CIFAR10, we report the results of best checkpoint and last checkpoint. The best results are marked using \textbf{boldface}.``\textbf{+LFRC}" represents the combination of our method with the previous method.}
    \label{white-ResNet-CIFAR10}
\end{table*}

%--------------------------------white wideResNet28-CIFAR100-----------------------------
\begin{table*}[ht]
    \begin{center}
    \begin{tabular}{c|ccccc|ccccc}
        \toprule
        \multirow{2}{*}{Method} & \multicolumn{5}{c|}{\textbf{Best Checkpoint}}        & \multicolumn{5}{c}{\textbf{Last Checkpoint}}         \\
                                & Clean & FGSM & PGD$_{20}$ & CW$_\infty$ &AA& Clean & FGSM & PGD$_{20}$ & CW$_\infty$ & AA \\ \noalign{\smallskip}\hline\noalign{\smallskip}
        AT                      &\textbf{60.67}&30.95&26.41&25.97&24.24&\textbf{59.26}&28.80&23.75&24.22&22.44    \\
        AT\textbf{+LFRC}        &59.12&\textbf{31.57}&\textbf{27.36}&\textbf{26.99}&\textbf{24.96}&57.84&\textbf{29.78}&\textbf{25.23}&\textbf{25.40}&\textbf{23.69}   \\ 
        \noalign{\smallskip}\hline\noalign{\smallskip}
        TRADES                  &\textbf{57.27}&33.62&30.93&27.81&26.78&\textbf{57.97}&31.80&27.97&26.87&25.79     \\
        TRADES\textbf{+LFRC}   &56.27&\textbf{33.91}&\textbf{31.77}&\textbf{28.47}&\textbf{27.38}&57.30&\textbf{33.54}&\textbf{30.62}&\textbf{27.80}&\textbf{26.87}    \\ \noalign{\smallskip}\hline\noalign{\smallskip}
        ALP         &59.84&30.77&26.11&25.00&23.83&59.77&30.75&26.06&24.89&23.75  \\
        ALP\textbf{+LFRC}       &\textbf{60.14}&\textbf{31.21}&\textbf{26.68}&\textbf{25.57}&\textbf{24.11}&\textbf{59.91}&\textbf{31.11}&\textbf{26.68}&\textbf{25.43}&\textbf{24.10}    \\
        \bottomrule
    \end{tabular}
    \end{center}
    \caption{The white-box robustness results(accuracy $(\%)$) of WideResNet28-10 on CIFAR100, we report the results of best checkpoint and last checkpoint. The best results are marked using \textbf{boldface}.``\textbf{+LFRC}" represents the combination of our method with the previous method.}
    \label{white-ResNet-CIFAR100}
\end{table*}

\section{Similarity matrices}
In \Cref{matrix}, we show more latent feature similarity matrices for natural and adversarial examples. 
It can be observed that in each pair of similarity matrices, the similarity matrices of natural examples are more compact (distinct bright blocks on the diagonal).

\begin{figure*}[t]
  \begin{center}
  \includegraphics[width=2.0\columnwidth]{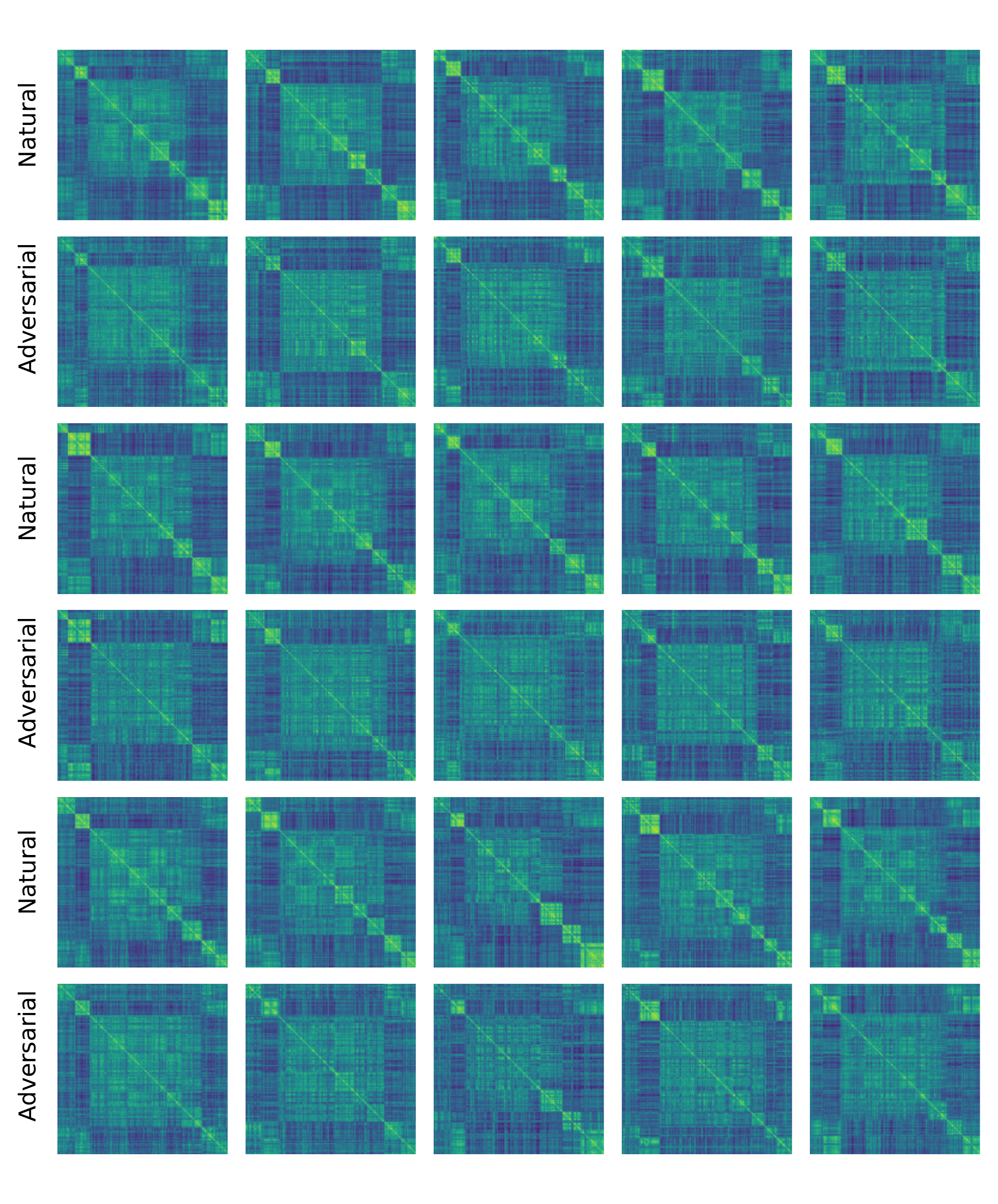}
  \end{center}
  \caption{More similarity matrices of the adversarial and natural examples on CIFAR10 test set for the ResNet18 network trained by Adversarial Training~\cite{madry2017towards}. The examples in each batch have been grouped by their true label for better visualization. The size of each similarity matrix is $256 \times 256$.}
  \label{matrix}
\end{figure*}

\end{document}